\documentclass{getwriting}
\usepackage{caption}
\usepackage{subcaption}
\usepackage{float}
\usepackage[figuresright]{rotating}
\usepackage{booktabs}

\usepackage{times}
\usepackage[bottom]{footmisc}

\title{Introduction to Machine Learning for Physicians:\\ A Survival Guide for Data Deluge}

\author[$\spadesuit$]{Ri\v{c}ards Marcinkevi\v{c}s, MSc \orcidlink{0000-0001-8901-5062}}
\author[$\spadesuit$]{Ece Ozkan, PhD \orcidlink{0000-0002-9889-6348}}
\author[$\spadesuit$]{Julia E. Vogt, PhD \orcidlink{0000-0002-6004-7770} \thanks{julia.vogt@inf.ethz.ch}}

\affil[$\spadesuit$]{\footnotesize Department of Computer Science, ETH Zurich, Zurich, Switzerland}

\begin{document}

\maketitle

\begin{abstract}
\noindent Many modern research fields increasingly rely on collecting and analysing massive, often unstructured, and unwieldy datasets. Consequently, there is growing interest in machine learning and artificial intelligence applications that can harness this `data deluge'. This broad nontechnical overview provides a gentle introduction to machine learning with a specific focus on medical and biological applications. We explain the common types of machine learning algorithms and typical tasks that can be solved, illustrating the basics with concrete examples from healthcare. Lastly, we provide an outlook on open challenges, limitations, and potential impacts of machine-learning-powered medicine.
\\\\
{\scriptsize \textbf{Keywords:} Machine Learning, Artificial Intelligence, Data Science, Precision Medicine, Predictive Medicine}
\end{abstract}


\emph{Machine learning} (ML) is a discipline emerging from computer science \cite{mlBishop2006} with close ties to statistics and applied mathematics. Its fundamental goal is the design of computer \mbox{programs \cite{mlMitchell1997},} or \emph{algorithms}, that learn to perform a certain task in an automated manner. Without explicit rules or knowledge, ML algorithms observe and possibly, interact with the surrounding world by the use of available data. Typically, as a result of learning, algorithms distil observations of complex phenomena into a general \emph{model} which summarises the \emph{patterns}, or regularities, discovered from the data.

Modern ML algorithms regularly break records achieving impressive performance at a wide range of tasks, \emph{e.g.} game playing \cite{alphagoSilver2017}, protein structure prediction \cite{alphafoldJumper2021}, searching for particles in high-energy physics \cite{searchingBaldi2014}, and forecasting precipitation \cite{precipitationRavuri2021}. The utility of machine learning methods for healthcare is apparent: it is often argued  that given vast amounts of heterogeneous data, our understanding of diseases, patient management and outcomes can be enriched with the insights from machine \mbox{learning \cite{mlandmedObermeyer2016,mlinmedRajkomar2019,mlinbiomedGoecks2020,mlinoncoCuocolo2020,introtomlGibbons2019,mlinepiRoth2018,mlinmedMay2021}.}

In this overview paper, we will provide a nontechnical introduction to the ML discipline aimed at a general audience with an affinity for biomedical applications. We will familiarise the reader with the common types of algorithms and typical tasks these algorithms can solve (Section~\ref{sec:whatIsML}) and illustrate these basic concepts by concrete examples of current machine learning applications in healthcare (Section~\ref{sec:ML_appl}). We will conclude with a discussion of the open challenges, limitations, and potential impact of machine-learning-powered medicine (Sections~\ref{sec:open_challenges}-\ref{sec:conclusion}).

\section{Machine Learning: An Overview}
\label{sec:whatIsML}

\paragraph{Why Now?}

Computer-based systems have become an integral part of modern hospitals \cite{compinmedSheridan2018} in numerous routine activities, including medical record keeping \cite{ehrJha2009}, imaging \cite{cadDoi2007}, and patient monitoring \cite{insulinpumpsUmpierrez2018}. Computer-aided diagnosis and treatment planning have been contemplated ever since the early days of computing. An iconic example is MYCIN --- an artificial-intelligence-based expert system developed in 1972 for diagnosing blood infections \cite{aiRussel2010,mycinBritannica2018}. In contrast to the purely data-driven machine learning perspective, MYCIN relied on about 450 hard-coded rules \cite{aiRussel2010} and logical reasoning to deduce a patient's diagnosis.

The machine learning approach is strikingly different: harnessing large and complex datasets, inducing rules or other types of patterns automatically, alleviating the need for tedious and costly knowledge engineering. With recent technological advances, we can now collect, store, and share health data at previously unprecedented scales \cite{bigdataShilo2020}, 
for instance, as of December 2021, The Cancer Imaging \mbox{Archive \cite{tciaClark2013}} contains almost 250,000 computerised tomography images \cite{tciaStats2021}. Such scales alone warrant the need for automated and data-driven methods in healthcare and clinical decision-making. 

\paragraph{Navigating through Concepts}

In the mediasphere and even academic literature, terms such as \emph{artificial intelligence}, \emph{machine learning}, \emph{deep learning}, and \emph{statistics} are sometimes used interchangeably  without drawing explicit borders \cite{mlvsaiKersting2018,mlvsstatsHarrell2021}. Below we provide a remark on the differences between these fields, which might be useful to a non-expert reader. 

\emph{Artificial intelligence} (AI) \cite{aiMcCarthy2007,aiRussel2010} tackles the most general problem of building \emph{intelligent} machines and is not restricted to a particular set of methods: both a simple expert system with a few hand-engineered rules and a logical inference engine \cite{mycinBritannica2018} or a deep artificial neural network playing the board game of Go \cite{alphagoSilver2017} could be seen as instances of artificial intelligence. On the other hand, ML is a subfield of AI \cite{aiRussel2010} focusing on the machines that learn from experience, namely, from the given data. \mbox{\emph{Deep learning} (DL) \cite{dlGoodfellow2016,dlSchmidhuber2015}} refers to an even smaller subset of machine learning methods: it studies deep neural networks (DNNs), a family of ML techniques which learn many layers of complex, highly nonlinear concepts directly from the raw data, \emph{e.g.} images, sound recordings, text, and videos. 

A well-informed reader might have noticed before that many ML methods rely on statistical reasoning and in some cases, ML and statistical models can be used for similar purposes \cite{statvsmlBzdok2018}. A simplified, stereotypical delineation between the two fields, sufficient for the current overview, is that classical statistical models are probabilistic, focus on \emph{inference}, and make strict structural assumptions; whereas ML methods offer an algorithmic solution to the \emph{prediction} problem allowing for very general and complex \mbox{relationships \cite{mlvsstatsHarrell2021,statvsmlBzdok2018}.}

\paragraph{Machine Learning Approaches and Tasks}

\begin{figure}[H]
    \centering
    \begin{subfigure}{1.0\linewidth}
        \centering
	\includegraphics[width=0.9\linewidth]{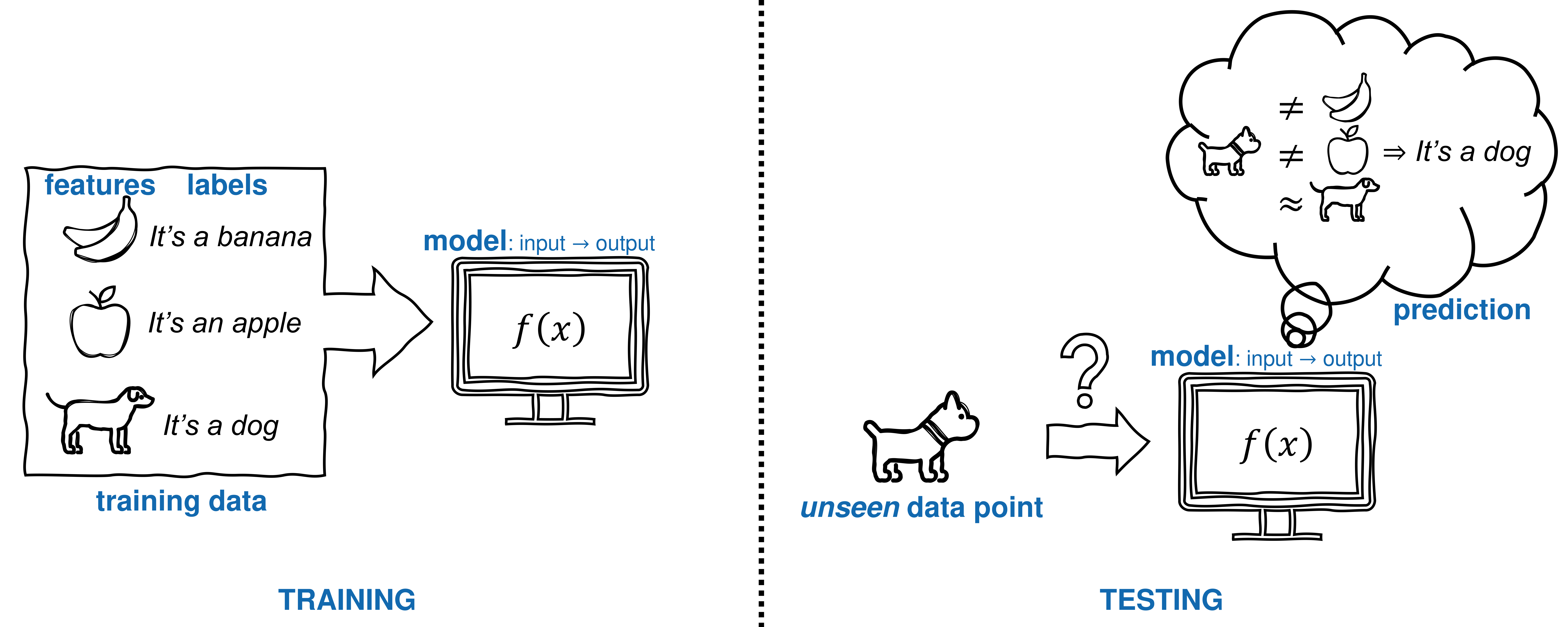}
        \caption{Supervised learning}
    \end{subfigure}
    \begin{subfigure}{0.75\linewidth}
        \centering
        \includegraphics[width=0.9\linewidth]{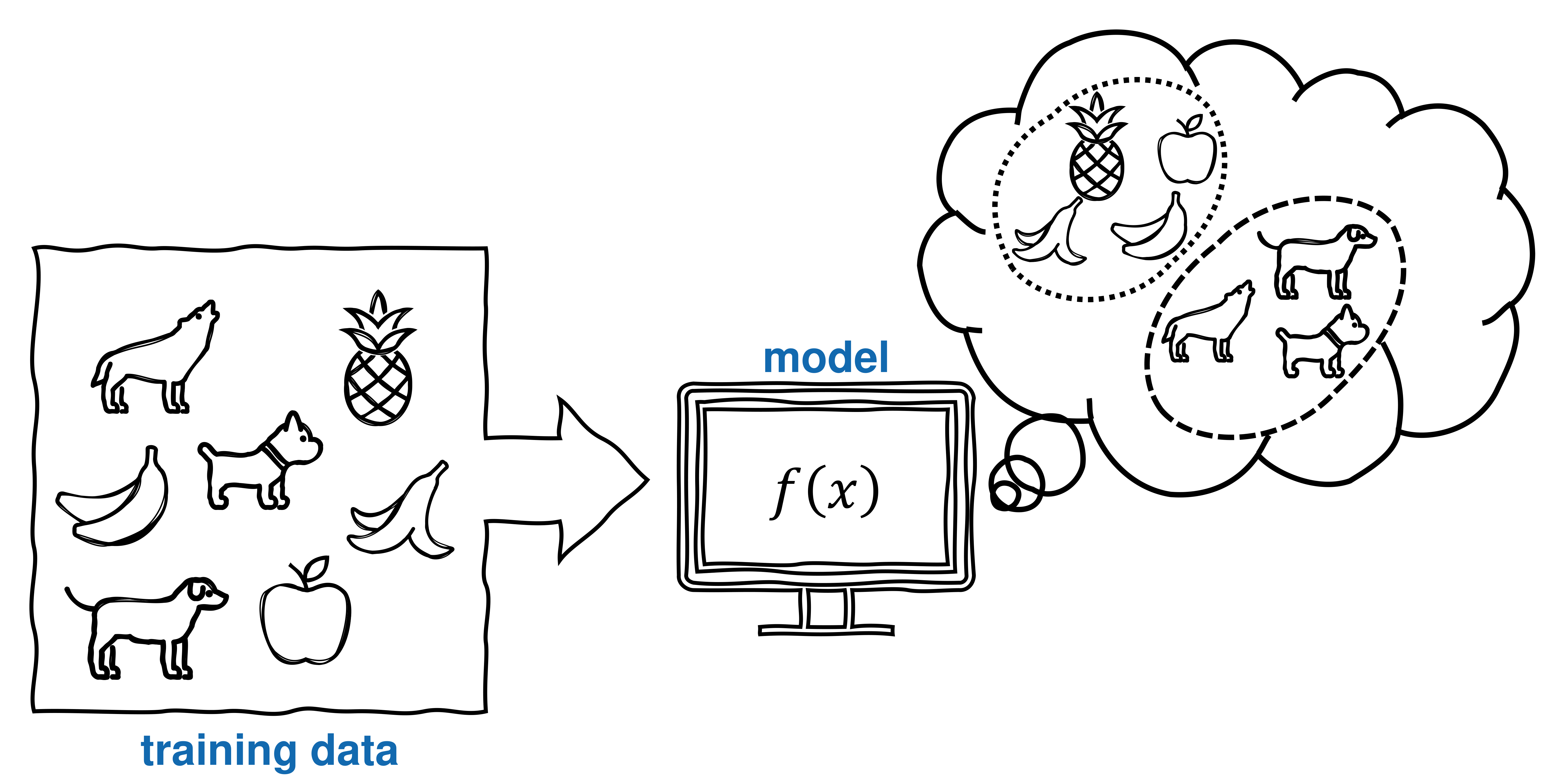}
        \caption{Unsupervised learning}
    \end{subfigure}
    \begin{subfigure}{0.75\linewidth}
        \centering
	\includegraphics[width=0.9\linewidth]{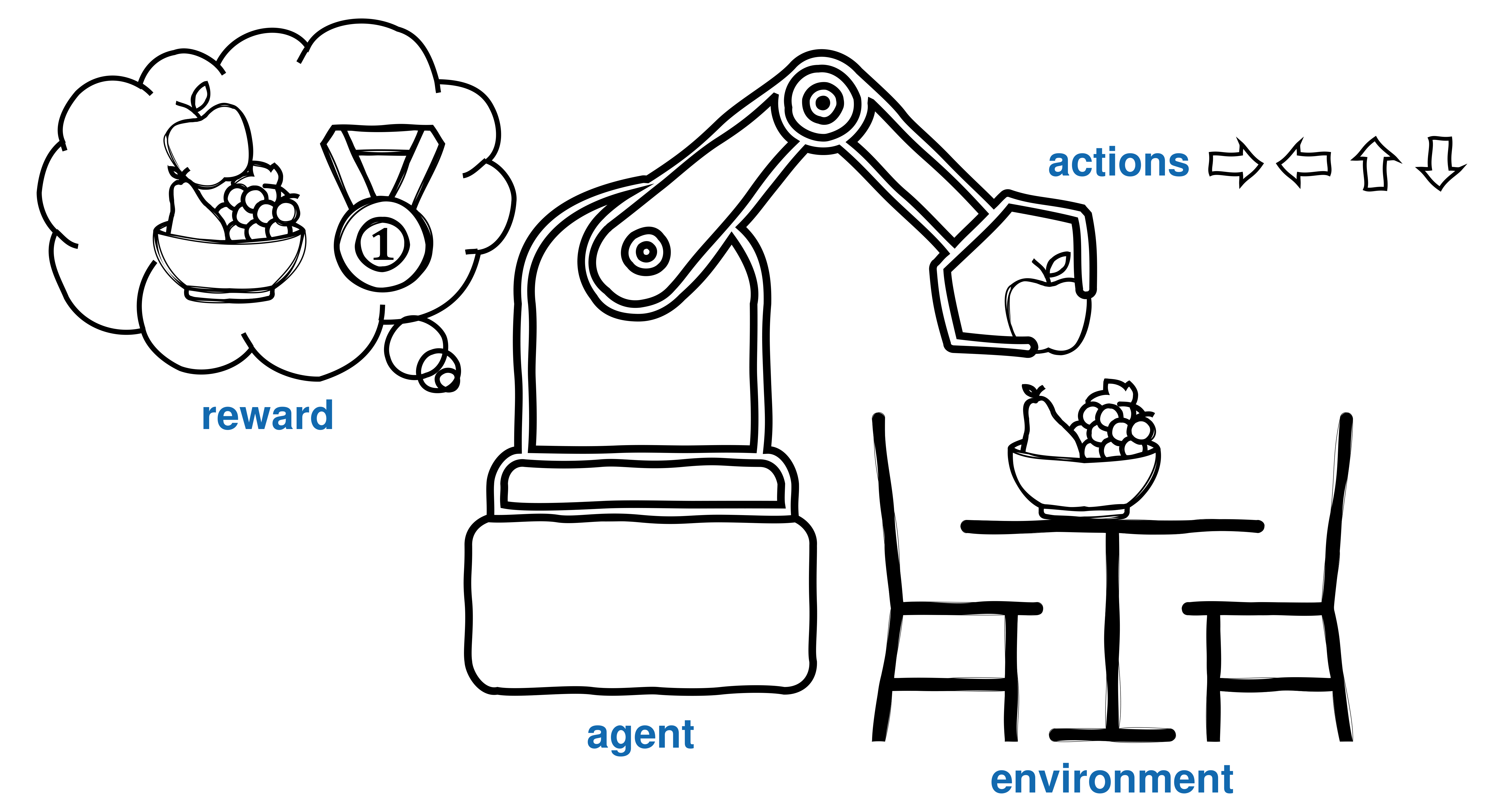}
        \caption{Reinforcement learning}
    \end{subfigure}
    \caption{Types of machine learning approaches: \mbox{(a) supervised}, \mbox{(b) unsupervised}, and \mbox{(c) reinforcement} learning. \mbox{(a) In supervised} learning, an ML model extracts predictive patterns from labelled data. \mbox{(b) In unsupervised} learning, a model discovers structure from unlabelled data. \mbox{(c) In reinforcement} learning, an agent interacts with the surrounding environment by sequentially performing actions to maximise received reward.  \label{fig:learning_types}}
\end{figure}

Three most common categories of machine learning methods are (\emph{i}) supervised learning, (\emph{ii}) unsupervised learning, and (\emph{iii}) reinforcement \mbox{learning \cite{mlBishop2006,mlMurphy2012} (Figure~\ref{fig:learning_types}).} These approaches have been tailored towards principally different problem types, which will be outlined below. Regardless of the method, ultimately, a model that \emph{generalises} well is desired, \emph{i.e.} a model with good performance at the considered task across as many different unobserved settings as possible.

In \emph{supervised learning} \cite{mlBishop2006,mlMurphy2012}, the goal is to learn a predictive relationship between a set of \emph{input variables}, also called \emph{features}, \emph{attributes}, or \emph{covariates}, and an \emph{output variable}, also called \emph{response}, \emph{target}, or \emph{label}. Following the example from Figure~\ref{fig:learning_types}(a), let us assume that we want to design a computer system to recognise objects, such as apples, bananas, or dogs from hand-drawn images. For this task, our features could be given by digital images stored on a personal computer, and labels could be verbal descriptions of the depicted objects, such as `\emph{It's a dog}', written down in a text file. Typically, we would use some \emph{learning algorithm} to extract predictive patterns from the observed \emph{training data}. The algorithm serves as a `recipe' for distilling the raw data into a sufficiently abstract and general \emph{model}. We would then apply the trained model to unseen \emph{test data} to predict unobserved labels, and evaluate its performance, for example, in terms of accuracy. The setting where labels come from a finite number of un-ordered categories, as in the hand-drawn image recognition example, is known as the \emph{classification} task; whereas in the \emph{regression} task, labels are usually real-valued. Note, that more generally, we could even learn relationships between input variables and \emph{multiple} differently-valued outputs.

\emph{Unsupervised learning} \cite{mlBishop2006,mlMurphy2012,unsupervisedJames2013} approach is not as well-defined as supervised learning and attempts to solve a more challenging open-ended problem: given only the input variables without labels, unsupervised learning algorithms typically seek to discover some `interesting' structure in the data. For instance, we might want to stratify our dataset into groups of similar observations --- this problem is known as \emph{clustering}. Following the hand-drawn image example above, a clustering algorithm would group images of similar objects together, \emph{e.g.} into groups of fruit and animals (Figure~\ref{fig:learning_types}(b)). \emph{Dimensionality reduction} is another typical unsupervised task where usually for visualisation purposes, we reduce the data to two or three informative dimensions by combining the input features. Although it might seem that supervised and unsupervised learning solve completely disjoint sets of problems, there is a whole spectrum of practical settings \mbox{in-between \cite{sslvanEngelen2019,wslZhou2017},} for example, when the output variable is partially missing because it is too expensive to measure for all of the subjects or when the output does not correspond to the exact target, which is impossible or unethical to measure. Techniques that deal with such settings fall under the categories of \mbox{\emph{semi-supervised} \cite{sslvanEngelen2019}} and \mbox{\emph{weakly supervised} \cite{wslZhou2017}} learning.

In their simplest forms, both supervised and unsupervised learning usually assume that the model is trained on a set of examples collected prior to learning and that model's predictions or decisions do not affect each other. \emph{Reinforcement learning} (RL) \cite{mlBishop2006,mlMurphy2012,rlSutton2018} ventures beyond these assumptions: the learning algorithm, in this context referred to as \emph{agent}, interacts with the surrounding \emph{environment} by observing it and performing sequences of \emph{actions} in order to maximise the occasionally obtained \emph{reward}. For example, a baby learning to walk and being praised by its parents or a robotic arm learning to place objects into a container (Figure~\ref{fig:learning_types}(c)) could be thought of as reinforcement learning scenarios. An important feature of this setting is the interactive and sequential nature of the learning process.

While the three approaches discussed above have become active research areas in their own right, in practice, real-world applications often require a combination of different methods and solving multiple tasks at a time. In the following, we will focus on biomedical and healthcare applications of ML supplementing our discussion with examples of the current research.  

\section{Machine Learning for Healthcare}\label{sec:ML_appl}

In healthcare, ML methods are usually leveraged to extract patterns that correlate with medical conditions. They are applied to healthcare records and other patient data to support clinicians by predicting diagnosis, management, and outcome in an automated manner. 

There are a plethora of data types in healthcare that ML algorithms can be applied to:
\begin{itemize}
    \item \textit{clinical data} from electronic health records (EHR) which contain demographics, laboratory test results, medication, allergies, vital signs, clinical reports \cite{deepMiotto2016,mimicJohnson2016},
    
    \item \textit{imaging data} from different modalities such as X-ray, retinal, mammography, dermoscopy, MR (magnetic resonance), CT (computerized tomography), and ultrasound (US) images and \\ \mbox{echocardiograms \cite{developmentGulshan2016,dermatologistEsteva2017,chexpertIrvin2019},} 
    
    \item \textit{sensors or mobile data} from wearable devices or sensors recorded and stored as time \mbox{series \cite{sleepSathyanarayana2016,wearableLonini2018},}
    
    \item \textit{omics data} which are complex and high-dimensional genomic, epigenomic, transcriptomic,\linebreak metabolomic, exposomic, and proteomic measurements \cite{predictingZhou2015,predictingAlipanahi2015}. 
\end{itemize} 
An advantage of ML methods over conventional statistical modelling is their flexibility and scalability in exploiting diverse and complex data types. 
Table~\ref{tab:applications} summarises a few selected application examples in terms of their domain, type of data, ML approach, and task.
Below we compile further recent works using various methods and types of data mentioned above. 

\paragraph{Supervised Learning}
Arguably, the most common ML approach in healthcare is supervised learning. 
Linear \cite{linregSanford2014} and logistic \cite{logregStoltzfus2011} regression models and decision trees \cite{decisionQuinlan1986} are simple, reliable, and effective supervised learning algorithms used on a variety of datasets to this day. Some examples include the classification of sagittal gait patterns in neurology \cite{applicationZhang2019}, prediction of diabetes severity \cite{comperativeKarun2019}, breast cancer survival prediction \cite{predictingUmer2008,decisionAzar2013}, diagnosis of acute appendicitis \cite{comparisonZorman2001} using clinical data, activity \mbox{classification \cite{activityParkka2006}} and fall detection in elderly people \cite{fallDiana2018} from wearable sensors, and prediction of interactions between target genes and \mbox{drugs \cite{gradientXuan2019}} using omics data.

However, simpler models often under-perform on data featuring nonlinear effects and interactions, and therefore, more flexible methods may be required, such as random forests \cite{rfBreiman2001} or gradient boosting machines \cite{gbFriedman2002}. Some of the applications of random forests are the prediction of the risk of developing hypertension \cite{interpretabiliyElshawi2019}, 
the severity and outcome of COVID-19 \cite{covidIwendi2020}, breast cancer recurrence \cite{explanationStrumbelj2010}, identification of various diseases, such as hepatitis, Parkinson's \cite{randomAlam2019,wearableLonini2018} or Type 2 Diabetes Mellitus \cite{machineZheng2017} using clinical and wearable sensors data. Furthermore, applications based on omics data also exist, \emph{e.g.} for improving hazard characterisation in microbial risk assessment \cite{improvingNjage2018} or identifying gene signatures for the diagnosis of \mbox{tuberculosis \cite{comparisonBobak2018}}. 

In medical images, the most commonly used models are based on DNNs, applied to a number of different modalities including 
\begin{itemize}
    \item  \textit{X-ray images} for predicting pneumonia \cite{transferRahman2020}, tuberculosis \cite{CheXaidRajpurkar2020} or other pathologies \cite{chexpertIrvin2019}, multi-organ segmentation \cite{automaticLi2019}, image registration \cite{cnnMaio2016}, pathology \mbox{localisation \cite{assessingArun2020,analyzingRajaraman2020,deepPark2020};} 
    \item \textit{retinal images} for predicting diabetic retinopathy grading \cite{developmentGulshan2016,usingSayres2019}, age-related macular \mbox{degeneration \cite{deepYan2020}} or retinopathy of prematurity \cite{automatedWorrall2016}, simultaneous segmentation of retinal anatomical structures, such as retinal vessel and optic disc \cite{deepManinis2016}, or pathologies, such as exudates, haemorrhages, microaneurysms, or retinal neovascularization \cite{retinalLam2018,simultaneousBadar2018};
    \item \textit{heart echocardiograms} for cardiac image segmentation \cite{segmentationYu2017,deepChen2020,cascadedXu2020}, identification of local cardiac structures and estimation of cardiac function \cite{fullyZhang2018,deepGhorbani2020,videoOuyang2020}, view classification \cite{realOstvik2019}, and the reduction of user variability in data acquisition \cite{automaticAbdi2017};
    \item \textit{mammography images} to improve breast cancer classification \cite{deepLi2019,robustLotter2021}, \\ lesion \mbox{localisation \cite{probabilisticErtosun2015,deepDhungel2017},} and risk assessment \cite{understandingMohamed2018,computerLi2018};
    \item \textit{dermoscopy images} for classifying skin cancer \cite{dermatologistEsteva2017,deepCodella2017,deepLiu2020}, skin lesion segmentation \cite{skinAlmasni2018}, or detection and localisation of cutaneous vasculature \cite{computerKharazmi2018};
    \item \textit{MR images} for brain tumour grading assessment \cite{automaticPereira2018}, classification of prostate cancer\cite{classificationSchelb2019}, age estimation \cite{ageHuang2017} or Alzheimer's disease detection \cite{novelIslam2017} from brain images, cardiac multi-structures segmentation \cite{deepBernard2018} or left ventricle segmentation in cardiac MRI \cite{combinedAvendi2016};
    \item \textit{CT images} for liver tumour assessment \cite{towardsCouteaux2019}, classification of brain images for diagnosis of Alzheimer's disease \cite{classificationGao2017}, haemorrhage detection \cite{radnetGrewal2018,deepChilamkurthy2018}, COVID-19 pneumonia classification \cite{multiAmyar2020}, abdominal \cite{automatedWeston2019} or urinary bladder \cite{urinaryCha2016} segmentation. 
\end{itemize}
There is accumulating evidence that DNNs have already led to improved accuracy for computer-aided applications in various medical imaging modalities. 
Moreover, interest and advances in deep learning are still developing rapidly in the ML for healthcare community, bringing the performance of ML methods to a level that is acceptable by the clinicians. 

\paragraph{Semi-supervised Learning}

The cost of acquiring raw medical images is often negligible in comparison to expert annotations, labels, or scores.
Thus, instead of using expensive supervised learning, many authors leverage semi-supervised learning (SSL) approaches \cite{sslvanEngelen2019}, which rely on unsupervised learning techniques to work with a mixture of labelled and unlabelled data. 
For medical images, the scarcity of labelled data is often a limitation for both segmentation and classification tasks. 
To this end, SSL has been leveraged for X-ray \cite{semiBortsova2019}, MR \cite{semiBai2017,ASDNetNie2018,shapeLi2020} and CT \cite{semiLuo2021} image, MS lesion \cite{semiBaur2017} and \mbox{gland \cite{deepZhang2017}} segmentation.
For the classification task, SSL methods have been applied for cardiac abnormality classification in chest X-rays \cite{semiMadani2018}, and skin lesion diagnosis \cite{semiLiu2020} or gastric disease identification from endoscopic images \cite{leveragingShang2019}. 
Pulmonary nodule detection in CT scans \cite{focalMixWang2020} using SSL is also possible. 

\paragraph{Unsupervised Learning}

Unsupervised learning methods further loosen the burden of manual annotation by exploiting unlabelled data without any supervision. 
It might be helpful to find interesting patterns or structure in the data, which can be, for example, used for anomaly detection. 
Various problems in the healthcare setting can be solved using unsupervised learning: disease clustering and patient subgrouping using \mbox{EHR \cite{unsupervisedWang2020},} fMRI time series \cite{variationalZhao2019} data or genomic makeup \cite{unsupvisedLopez2018}, genomic \mbox{segmentation \cite{unsupervisedHoffman2012},} dimensionality reduction prior to anomaly \mbox{detection \cite{deepZong2018}} or classification of mammography \mbox{images \cite{paretoTaghanaki2017}.} 
Unsupervised learning has been applied to various imaging modalities, \emph{e.g.} X-ray image retrieval \cite{convolutionalAhn2019}, blood vessel segmentation in retinal images \cite{bloodZhang2015}, breast density segmentation and scoring in \mbox{mammography \cite{unsupervisedKallenberg2016},} probabilistic atlas-based segmentation \cite{unsupervisedDalca2019} and denoising of contrast-enhanced sequences in MR \mbox{images \cite{ensembleBenou2017},} and lung nodule characterisation in CT \mbox{images \cite{lungHussein2019}.} 

\paragraph{Reinforcement Learning}
In RL, an agent learns from new experiences through a continuous trial-and-error process \cite{applicationsMahmud2018}.
This approach has been used in different medical scenarios, \emph{e.g.} for optimal and individualised treatment strategies for HIV infected patients \cite{clinicalErnst2006}, for cancer trials \cite{reinforcementZhao2009}, and for sepsis patients \cite{individualizedSaria2018}, forecasting the risk of diabetes \cite{forecastingZohora2020}, segmenting transrectal US images to estimate location and volume of the prostate \cite{applicationSahba2008}, surgical decision-making \cite{artificialKomorowski2018}, predicting CpG islands in the human genome \cite{particleChuang2011}, increasing the accuracy of biological sequence annotation \cite{reinforcementRalha2010}. 

\begin{sidewaystable}[ph!]
    \caption{A few selected healthcare application examples summarised in terms of the application domain, type of data, ML approach, and task.\label{tab:applications}}
    {
    \begin{center}
    \begin{tabular}{cccp{32mm}p{90mm}c}
        \toprule
        \textbf{Application Domain} & \textbf{Data Type} & \textbf{ML Approach} & \textbf{ML Task} &  \textbf{Description} & \textbf{Ref.}\\ 
        & \\
        \hline
        Ophthalmology & Imaging & Supervised & Classification & Predicting diabetic retinopathy grading using retinal fundus photographs &  \cite{developmentGulshan2016} \\
        \hline
        Dermatology & Imaging & Supervised & Classification  & Classifying skin lesions comprising 2,032 different diseases &  \cite{dermatologistEsteva2017} \\
        \hline
        Cardiology & Imaging & Supervised & Segmentation  & Segmentation of fetal left ventricle in echocardiographic sequences &  \cite{segmentationYu2017} \\
        \hline
        Infectiology & Clinical & Supervised & Classification  & Predicting the severity and outcome of COVID-19 &  \cite{covidIwendi2020} \\
        & \\
        \hline
        Genetics & Omics & Unsupervised & Segmentation & Semi-automated genomic annotation &  \cite{unsupervisedHoffman2012} \\
        & \\
        \hline
        Oncology & Imaging & Unsupervised & Denoising & Spatio-temporal denoising of contrast-\newline enhanced MRI sequences &  \cite{ensembleBenou2017} \\ 
        \hline
        Endocrinology & Clinical & Unsupervised & Dim. reduction, \newline anomaly detection & Anomaly detection for several public benchmark datasets including thyroid dataset & \cite{deepZong2018} \\
        \hline
        Urology & Imaging & RL & Segmentation & Segmenting transrectal US images to estimate location and volume of the prostate & \cite{applicationSahba2008} \\
        \hline
        Oncology & Clinical & RL & Regression & Individualised treatment strategies for advanced metastatic stage lung cancer &  \cite{reinforcementZhao2009} \\
        \hline
        ICU & Clinical & RL & Classification & Personalised patient-centred decision-making &  \cite{artificialKomorowski2018} \\
        & \\
        \hline
    \end{tabular}
    \end{center}
    }
\end{sidewaystable}

\bigskip

The rapid progress in ML triggered considerable interest and created numerous opportunities for data-driven applications in healthcare, which in the future might lead to advancements in clinical practice like semi-automated and more accurate diagnosis or the development of novel and more personalised treatment strategies. 
Despite these reasons for cautious optimism, plenty of under-explored opportunities, unaddressed limitations and concerns exist. 
In the following, we elaborate on some open problems.

\section{Limitations, Challenges, and Opportunities}\label{sec:open_challenges}

While it is undoubted that ML can be instrumental in developing data-driven decision support systems, the use of machine learning and promises associated with it have come under \mbox{scrutiny \cite{mlandmedObermeyer2016,precisionmedWilkinson2020,predmodelsChen2020}.} Machine learning is not a panacean magic crystal ball: it relies on large amounts of high-quality, representative, and unbiased data \cite{mlandmedObermeyer2016} and is not a substitute for rigorous study and data system design or causal \mbox{inference \cite{precisionmedWilkinson2020}.} Prior to applying ML methods to health data, we should be critical and ask ourselves  whether there actually exists use cases for our ML model \cite{predmodelsChen2020}, whether we possess a sufficiently large dataset representative of the population of interest, whether the features we will use are commonly available in practice, whether we would be able to validate the model externally, whether our model will be accepted by its target \mbox{users \cite{acceptGardner1994,doctorsLipton2017}.} Some of the recent lines of work attempt to alleviate and address these and similar concerns. Below we discuss a few open challenges and areas of active research which could be of interest to both ML methodologists and practitioners working in healthcare.  

\paragraph{Lack of Data, Labels, and Annotations}
Data scarcity is a very important problem since data lie at the heart of any ML project. 
For most of the applications, in addition to data collection, the annotation is an expensive task. Furthermore, medical data are personal and accessing and labelling it has its own challenges. 
Moreover in case of rare diseases, class imbalance makes the use of ML algorithms more challenging. 
To overcome these problems, different terms for learning are proposed which do not entirely depend on manually labelled data: \emph{self-supervised learning}, where the model leverages the underlying structure of the data \cite{selfslLeCun2021}, \emph{semi-supervised learning} using a small amount of labelled combined with a large amount of unlabelled data during training \cite{sslvanEngelen2019}, \emph{weakly-supervised learning} using the supervision of noisy labels \cite{wslZhou2017}, and \emph{few-shot learning} for generalising from a small amount of labelled data without using any unlabelled data \cite{fslWang2021}. 

\paragraph{Learning across Domains, Tasks, and Modalities}
ML methods generally assume that the training and test sets feature similar patterns and relationships, which usually does not hold for healthcare applications. 
\emph{Domain adaptation} is a promising solution and gets an increasing attention in recent \mbox{years \cite{domainsGuan2021}.} 
Furthermore, trained models usually suffer from overspecialising at individual tasks which tend to not generalise \mbox{either \cite{pathwaysDean2021}.} 
As a response, \emph{multi-task learning} was proposed, which is inspired by human learning and proposes to use shared concepts to extract the common ideas among a collection of related tasks \cite{multitaskCaruana1997,Crawshaw2020MultiTaskLW}. 
Moreover, clinicians typically make use of multiple data modalities in their decision-making, including imaging, time series data, such as ECG signals, clinical data, such as lab results, and non-structured data, such as clinical notes. 
Combining data from different modalities lies at the centre of \mbox{\emph{multimodal learning} \cite{multimodalBaltrusaitis2018,multimodalBaltrusaitis2019}.}

\paragraph{Data Sharing, Privacy, and Security}
Despite the spectacular advances in the ML domain, there is a concern regarding privacy and security in the proposed data-driven methods. This is traditionally quite a challenging issue in the healthcare setting due to the fact that ML models need to work with personal \mbox{information \cite{securemlhealthQayyum2021}.} 
The need for patient privacy while implementing ML methods using large datasets triggers the urge for automated models which respect privacy and security. The issue has recently been picked up by official authorities, \emph{e.g.} in the EU the General Data Protection Regulation (GDPR) was \mbox{implemented \cite{europeanGoodman2016,gdprBovenberg2020}.} 
This decision accelerated the field of secure and privacy-preserving ML research to bridge the gap between data protection and utilisation for clinical routine  \cite{securemlKaissis2020,securemlZhang2019}.

\paragraph{Interpretable and Explainable Machine Learning}

Machine learning models are increasingly incorporated into high-stakes decision-making \cite{doctorsLipton2017,stopblackboxRudin2019}, including healthcare \cite{triageHao2020}. With methodological advances and empirical success of deep learning, ML models have become even more performant, yet larger and more complex. As a response, there is a surge of interest in designing ML systems that are transparent and trustworthy. \emph{Interpretable} \cite{towardsDoshiVelez2017,mythosLipton2018,stopblackboxRudin2019} and \emph{explainable} \cite{explainableSamek2019} machine learning typically refer to models that are either directly comprehensible or can explain and present their decisions \emph{post hoc} in terms understandable to the target user. Many interesting research questions originate from this line of work, pertinent to healthcare as well. For instance, does ML even need to be interpretable or explainable, what is a `good' interpretation/explanation in the considered application, how can we make model's interpretations/explanations insightful and/or actionable?

\paragraph{Fair Machine Learning}

The fact that ML methods are data-driven does not necessarily make their decisions fair, ethical, or moral \cite{fairmlBarocas2019}. Datasets and ML models are products of the larger sociotechnical system we live in and inevitably reflect the state of the society with all of its disparities \cite{aibiasesZou2018}. \mbox{\emph{Fairness} \cite{fairmlBarocas2019,fairmlhealthRajkomar2018}} has become an important principle for machine-learning-based algorithmic decision-making and given rise to many technical challenges, such as mitigating demographic disparities captured by ML models \cite{aif360Bellamy2018}, documenting and reporting ML models in a transparent manner \cite{modelcardsMitchell2019}, and documenting datasets, their contents, characteristics, and intended uses \cite{datasheetsGebru2021}. Many of these considerations are equally applicable to ML for healthcare \cite{fairmlhealthRajkomar2018} where personalisation based on sensitive information could sometimes lead to undesirable disparate outcomes \cite{fairmlBarocas2019}.

\paragraph{Towards Causally-informed Models}

Many clinical and scientific research questions inherently require causal reasoning \cite{causalityPearl1995}, \emph{e.g.} treatment effect estimation or counterfactual outcome prediction. The `big data' alone are meaningless, unless researchers are equipped with adequate tools that actually can address their questions \cite{causalRaita2021}. \emph{Causality} \cite{causalityPearl2009} is a field of study which formalises our reasoning about cause-and-effect relationships. In the recent years, it has been argued that many challenging open ML problems are closely related to causality and the inability of the na\"{i}ve purely data-driven models to reason \mbox{causally \cite{causalmlScholkopf2019,causalScholkopf2021,causalmlPearl2019}.} This realisation is an important step towards data collection and ML model development that are informed by the causal perspective. Naturally, causal considerations matter for healthcare applications of ML as well: causally-informed ML models tend to be more stable and avoid the pitfalls of a purely predictive approach. For example, when analysing medical imaging \mbox{data \cite{Castro2020}} with highly heterogeneous anatomical and acquisition conditions, causality could help mitigate irrelevant correlations picked up by a predictive model due to confounding in the training data. 

\section{Conclusion: Quo Vadis?}\label{sec:conclusion}

In this overview, we have introduced the basics of machine learning, types of tasks it is able to tackle, explained its relationships with other well-established fields, and illustrated its utility in healthcare with a variety of applied research. We have seen that opportunities for application are numerous: ML offers an algorithmic solution to the automated analysis of complex and unwieldy datasets. 

It is beyond doubt that ML and AI hold promises of automating routine clinical tasks \cite{automationJamieson2019}, reducing \mbox{costs \cite{costsNgiam2019},} and improving healthcare access and quality \cite{ml4hSarkar2020}, especially, in developing economies. Some researchers have even optimistically proclaimed that deep learning will replace human specialists altogether, for example, in radiology \cite{radiologyHinton2016,chexpertIrvin2019,radiologySabater2021}. Not surprisingly, their bold predictions have not come to pass yet, and a wide-spread, clinically meaningful adoption of ML `in the wild' still stands as an ambitious task for both the research and industry. 

The next decades will show if and how these promises are delivered. Likely, it is by supporting, complementing, and relieving healthcare professionals of tedious routine (and not by replacing them) that ML and AI will `make the difference'. To achieve this kind of harmony, we need an interdisciplinary collaboration among healthcare specialists, ML practitioners, and methodologists \cite{doctorsLipton2017}. To facilitate a dialogue on equal terms, we should improve the overall digital, machine learning, and statistical literacy among medical students \cite{mededucJames2021,studierendenBonvin2022} and disseminate subject matter knowledge among machine learning practitioners \cite{potTibshirani2021}. Lastly, to implement this ambitious project in practice, we have to develop regulatory frameworks and economic incentives \cite{mlinmedRajkomar2019}, establish large-scale data infrastructures \cite{datainfPanch2019}, and collect high-quality representative datasets \cite{datacentricWhang2021}. While all of this requires a considerable initial investment, the machine-learning-powered medicine is a worthy cause.

\bibliographystyle{ieeetr}
\bibliography{getwriting}

\end{document}